\RequirePackage{fix-cm}
\documentclass[twocolumn]{svjour3}          
\smartqed  
\usepackage{cite}
\usepackage{hyperref}
\usepackage{graphicx}
\usepackage{booktabs}
\usepackage{longtable}
\usepackage{multirow}
\usepackage{nccmath}
\usepackage{amsmath,amssymb}
\usepackage{footnote}
\DeclareMathOperator{\CRF}{CRF}
\DeclareMathOperator{\BiGRU}{BiGRU}
\DeclareMathOperator{\ALSA}{ALSA}%
\begin{document}

\title{Improving Aspect-Level Sentiment Analysis with Aspect~Extraction
}


\author{
Navonil Majumder$^\ast$ \and
Rishabh Bhardwaj$^\ast$ \and
Soujanya Poria$^\dagger$ \and
Amir Zadeh \and
Alexander~Gelbukh \and
Amir Hussain \and
Louis-Philippe Morency
}


\institute{N. Majumder and A. Gelbukh \at
              CIC, Instituto Polit\'enico Nacional, Mexico\\
              \email{navo@nlp.cic.ipn.mx, gelbukh@cic.ipn.mx}
           \and
           R. Bhardwaj and S. Poria \at
           Singapore Institute of Technology and Design, Singapore\\
           \email{rishabh\_bhardwaj@mymail.sutd.edu.sg, sporia@sutd.edu.sg}
           \and
           A. Hussain \at
           Edinburgh Napier University, UK\\
           \email{A.Hussain@napier.ac.uk}
           A. Zadeh and L.-P. Morency \at
           Carnegie Mellon University, USA\\
           \email{abagherz@andrew.cmu.edu, morency@cs.cmu.edu}
           \and
           $^\ast$ denotes equal contribution.\\
           $^\dagger$ denotes corresponding author.
}

\date{Received: date / Accepted: date}

\maketitle

\begin{abstract}

Aspect-based sentiment analysis (ABSA), a popular research area in NLP has two distinct parts --- aspect extraction (AE) and labeling the aspects with sentiment polarity (ALSA). Although distinct, these two tasks are highly correlated. The work primarily hypothesize that transferring knowledge from a pre-trained AE model can benefit the performance of ALSA models. Based on this hypothesis, word embeddings are obtained during AE and subsequently, feed that to the ALSA model. Empirically, this work show that the added information significantly improves the performance of three different baseline ALSA models on two distinct domains. This improvement also translates well across domains between AE and ALSA tasks.

\end{abstract}

\section{Introduction}
\label{intro}
Owing to the wide proliferation of smart-devices and internet across the world, opinion sharing over the internet has become a norm in modern society. This has given rise to various social-media platforms like Facebook, Twitter, Reddit, YouTube, where huge quantity of opinionated textual data on myriad of topics is being shared everyday. This is lucrative to large companies as they can use this data to perform market research, feedback gathering, risk assessment, marketing. These tasks have large and critical implications on their revenue, resource allocation, investments, and the company as a whole. As such, making sense of such huge volume of data warrants scalable and effective opinion mining systems. It is often convenient and useful to view this data in terms of sentiment, which requires
sentiment analysis algorithms~\cite{mohammad2013nrc, ruder2016hierarchical}. These algorithms operate at sentence level. However, it is often necessary to extract sentiment of individual topics/aspects within a sentence. As an example, if a company bring a new laptop to the market, they would be interested in user feedback on its various aspects, like display, battery life, keyboard, etc. Extracting such fine-grained sentiment calls for aspect-based sentiment analysis (ABSA) algorithms.

ABSA consists of two stages --- aspect extraction (AE) and aspect-level sentiment analysis (ALSA). Aspect extraction deals with identifying different aspects mentioned within a given sentence. One of the prominent approaches to aspect extraction has been dependency parse-tree-based~\cite{Qiu:2011:OWE:1970420.1970422,ae-poria-rule-based}. However, recently several neural network-based methods have been devised~\cite{shu2017lifelong,wang2016recursive}. On the other hand, ALSA determines the sentiment of the extracted aspects within given sentence. \cite{wang2016attention,ma2017interactive} proposed few of the neural network-based method for ALSA.

The works in \cite{shu-etal-2017-lifelong,wang-etal-2016-recursive} argue that modelling relationship
between words is key to effective aspect extraction. As such, we assume that any AE method
would learn these relationships. This is true in case of aspect-level sentiment analysis
also, as the ALSA algorithm needs to make correct association between target aspect and its
corresponding sentiment-bearing word. This becomes even more crucial in case of sentences
having multiple aspects. If an ALSA model could get information on the relationship between the words
in the sentence externally, then we believe it would use this information to identify the relevant
words for sentiment classification of the target aspect, which can improve the
classification performance. In other words, load sharing between AE and ALSA could improve the overall performance of ALSA. This inspires our
hypothesis that the relationships between the words learnt during AE can aid ALSA to
perform better than on its own. To this end, we learn word embeddings from AE and use this
in ALSA, a form of transfer learning. The hypothesis of the relatedness of AE and ALSA tasks has been considered by other works~\cite{luo2019doer}. But these works perform AE and ALSA tasks consecutively which causes ALSA performance drop by a large margin because the performance of ALSA depends on AE performance.

In this paper, we adopt bidirectional long short-term memory (BiLSTM) and conditional 
random field (CRF)-based sequence tagging 
method~\cite{DBLP:journals/corr/HuangXY15} for aspect extraction. Instead of BiLSTM, we feed
the sentence to a bidirectional gated recurrent unit (BiGRU) layer (see Section \ref{sec:ae-model}) that propagates contextual information among the words. This also
establishes relationship among the words. The output of this BiGRU is fed to a CRF layer for
labeling each word if it is part of an aspect term. We posit that after training such model,
the BiGRU output corresponding to each input word would contain information on all the
related words. As such, we pass this BiGRU output as auxiliary word embedding to the ALSA
model.

To demonstrate generality, we show efficacy of our approach on three different ALSA models,
namely TC-LSTM~\cite{tang-etal-2016-effective}, ATAE~\cite{wang-etal-2016-attention},
and IAN~\cite{ma2017interactive}. The embeddings learnt from AE are concatenated to the input words embeddings (GloVe) of ALSA models. In Section \ref{sec:results}, we show that adding
this extra information to ALSA results in significant performance improvement.

The rest of this paper is structured as follows --- Section \ref{sec:related-works} mentions 
various prominent methods of AE, ALSA, and transfer learning; Section \ref{sec:model} describes our AE to ALSA transfer learning scheme; Section \ref{sec:experimental-settings} the various
experiments we performed; Section \ref{sec:results} reports the results of our experiments and provides interpretation and analysis of those results; finally, Section \ref{sec:conclusions}
makes concluding remark by indicating the contributions of this paper and potential future directions.

\section{Related Works}
\label{sec:related-works}

Aspect-level sentiment classification is a fine-grained text classification problem. Various works~\cite{kaji2007building, rao2009semi, perez2012learning, caruana1997multitask, rana2016aspect, singh2013sentiment, steinberger2014aspect} have been done on detecting sentence-level polarity with handcrafted features. \sloppy{Recent advances introduced neural network-based approaches~\cite{socher2011semi, dong2014adaptive, mohammad2013nrc, ruder2016hierarchical, chen2020multi, nandal2020machine, halim2019efficient, shams2020lisa, halim2017profiling}}. Despite such approaches yielding promising results, these do not
cater to sentences expressing emotions on multiple topics.

The work in \cite{tang2015effective} uses two target-dependent LSTMs to model left and right contexts of the target, including the target string. To focus on the relevant parts of a sentence, \cite{wang2016attention} proposed an attention-based LSTM. IAN~\cite{ma2017interactive} generates the context and target representations by modelling interaction between them using attention. 

Aspect extraction is a crucial task in sentiment analysis and opinion mining~\cite{liu2012sentiment, pontiki2016semeval, angelidis2018summarizing}. \cite{hu2004mining} presents a distinction between kinds of aspects, i.e., explicit and implicit. \cite{popescu2007extracting, blair2008building} improved their method as it dealt with explicit aspects. Recently, deep neural networks (DNN) based approaches have outperformed traditional or rule-based approaches~\cite{poria2016aspect, zhang2018deep}. Deep Neural networks learn better representations to produce neural topic models which are more coherent topic  of text than previous approaches such as Latent Dirichlet Allocation (LDA)~\cite{he2017unsupervised, srivastava2017autoencoding}.

In recent years, transfer learning methods have brought significant improvements to wide range of NLP tasks by following learning in isolation paradigm~\cite{ruder2019transfer}. \cite{mikolov2013distributed} first introduced context-independent distributed word representations. Modern approaches learn sentence, document, or word representations that are context sensitive \cite{le2014distributed, conneau2017supervised, mccann2017learned, peters2018deep}. Moreover, most major improvements on named entity recognition (NER) involve some form of transfer learning--or auxiliary task--based self-supervised learning~\cite{ando2005framework}, clustering phrases~\cite{lin2009phrase}, pretrained language models and embeddings~\cite{peters2017semi, akbik2018contextual, baevski2019cloze}.

Although, transfer learning approaches have been used widely in NLP, for aspect-level sentiment classification knowledge transfer from aspect extraction has remained unexplored. To the best of our knowledge, our work is the first that introduces transfer learning for aspect-based sentiment analysis.

\section{Transfer Learning for Aspect-Based~Sentiment~Analysis}
\label{sec:model}

Aspect-based sentiment analysis (ABSA) usually consists of two tasks
---  aspect extraction (AE) and aspect-level sentiment analysis
(ALSA). Naturally, AE is performed prior to ALSA.

\subsection{Constituent Task Definitions}

\paragraph{Aspect Extraction (AE) ---} Given a sentence $S=[w_1,w_2,\dots,w_n]$ consisting of $n$ words,
the task is to label each word $w_i$ with one of the three labels $B,I,$ and $O$.
$B$ and $I$ represent the initial and non-initial words of constituent aspect terms,
respectively. Whereas, $O$ represents all the words that are not part of any aspect term.

\paragraph{Aspect-Level Sentiment Analysis (ALSA) ---} Upon aspect extraction (AE), we get
$m$ aspect terms $A_1, A_2,\dots, 
A_m$. Each $A_i$ consists of a convex subsequence of words in $S$ with corresponding 
sequence
of AE labels with regular expression pattern $BI*$. With all this information, the task 
is to classify each $A_i$ with appropriate sentiment polarity (\emph{positive}, 
\emph{negative}, or \emph{neutral}). 

\subsection{Transfer Learning Hypothesis}
\begin{figure*}[ht!]
    \centering
    \includegraphics[width=0.75\linewidth]{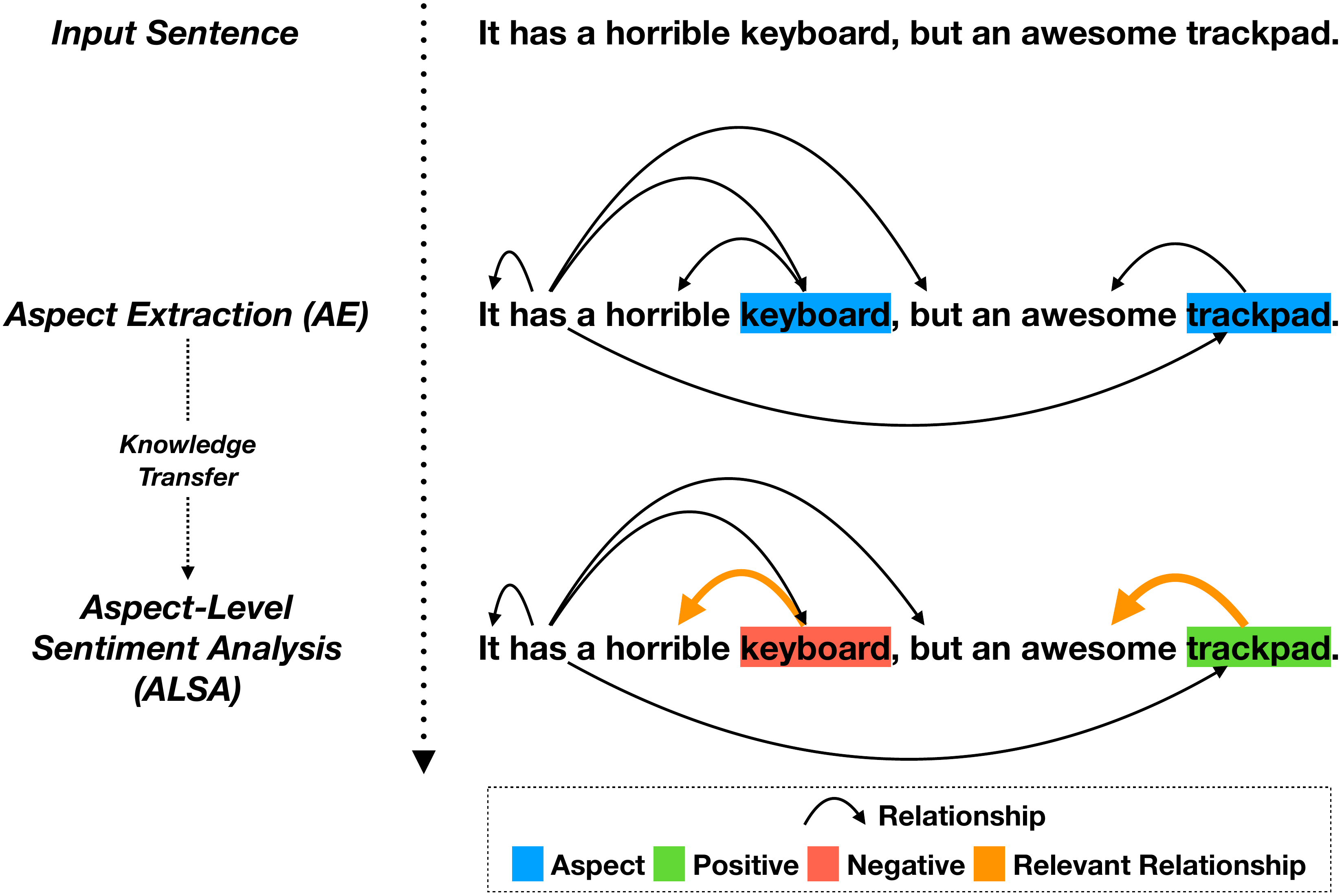}
    \caption{Illustration of hypothesized knowledge transfer from AE to ALSA; AE identifies the 
    aspects, along with learning dependencies; ALSA learns to identify
    relevant dependent words for classification, using the learnt dependencies from AE.}
    \label{fig:TL_example}
\end{figure*}

Our hypothesis is that the task of AE can aid the task of ALSA by providing ALSA with
the syntactic and semantic features learnt by AE. We posit that AE model learns
relationship between aspect terms and surrounding context words particularly well,
as it is required to do so to correctly identify the aspect words. Our supposition is 
supported by the works of 
\cite{shu-etal-2017-lifelong,PORIA201642,wang-etal-2016-recursive}, where dependency 
relations play a major role in AE. Since, some of the dependent words are bound to carry
sentimental information on aspect terms, this makes the job of ALSA model easier by 
providing it with information on the dependent contextual words. Now all, the ALSA model 
has to do is to learn the semantics of the relevant dependent words to make correct 
sentiment classification of the target aspect. \ref{fig:TL_example} illustrates this 
reasoning. 

Without the aid of AE model, ALSA model has to implicitly learn these dependency
relations among words, along with explicitly learning to assign sentiment labels. We believe the 
explicit learning overshadows the implicit learning. This is where the AE model compensates. 

\subsection{Aspect-Extraction Model}
\label{sec:ae-model}

Following the transfer learning hypothesis, we first train an AE model to learn the dependencies among the constituent words in sentence $S=[w_1, ..., w_n]$, where each word
$w_i\in \mathbb{R}^d$ is represented by a 300-dimensional pretrained GloVe\footnote{\url{https://nlp.stanford.edu/projects/glove/}}~\cite{pennington2014glove} embedding.
Since, AE is fundamentally a sequence tagging problem, we choose the BiLSTM-CRF-based sequence tagging method by \cite{DBLP:journals/corr/HuangXY15} as the
basis for our AE model. We simply replace the BiLSTM~\cite{hochreiter1997long} with bidirectional gated recurrent unit (BiGRU)~\cite{DBLP:journals/corr/ChungGCB14}
--- as GRUs have fewer parameters and yields similar performance to LSTM in our experiments --- 
and keep the rest unchanged.

Bidirectional RNN-based structure, like BiGRU, allows modelling 
relationships among elements of a sequence, both backward and forward along the sequence.
This property makes BiGRU an ideal candidate for learning relationships among
words in a sentence, which is essential for accurate aspect extraction (AE). Further, 
these learnt dependencies could presumably aid ALSA in our transfer learning setup. Thus,
the sequence of words in sentence $S\in \mathbb{R}^{n\times d}$ are fed to a BiGRU, named
$\BiGRU_{AE}$, of output
size $D_T$ to generate relationship-aware word representations in $S_T\in \mathbb{R}^{n\times D_T}$:
\begin{flalign}
    S_T=\BiGRU_{AE}(S),
    \label{eq:AE_1}
\end{flalign}
where $S_T=[v_1,v_2,\dots,v_n]$ and $v_i\in \mathbb{R}^{D_T}$ corresponds to relationship-aware representation of word $w_i$.

Since, there is a strong dependency between two consecutive words and their corresponding
aspect-identifying labels, conditional random field (CRF) is used to jointly model these
two dependencies. As such, the relationship-aware word representations in $S_T$ are fed to
CRF for the classification of the membership of the words ($C$) in an aspect term:
\begin{flalign}
    C=\CRF(S_T),
\end{flalign}
where $C\in \{B,I,O\}^n$.


\subsection{Knowledge Transfer from AE to ALSA}

Given a sentence $S$ having $n$ words that are represented by some vector of fixed length
and $m$ aspects, $A_1, A_2,\dots,A_m$, an ALSA model $\ALSA_{\ast}$ ($\ast$ represents any arbitrary ALSA algorithm like IAN~\cite{ma2017interactive}, TC-LSTM~\cite{tang-etal-2016-effective}, etc) outputs sentiment
label $l_i$ for each aspect $A_i$:
\begin{flalign}
    l_i&=\ALSA_{\ast}(S, A_i), \label{eq:alsa1}\\
    A_i&=S[h_i:k_i], \text{where }0\leq h_i\leq k_i \leq n, \label{eq:alsa2}
\end{flalign}
where $h_i, k_i$ are the indices of first and last word of aspect $A_i$ in $S$,
respectively.

We first train the AE model in Section \ref{sec:ae-model}. We assume that the output of trained $\BiGRU_{AE}$, the 
relationship-aware word representations ($S_T$), have the relationships among the
constituent words in $S$ encoded in them. We intend to leverage this learnt knowledge from AE in
ALSA. To this end, the word embeddings ($w_i$) of each sentence for ALSA are concatenated by their
corresponding representation in $S_T$; we represent this word-wise concatenation of representations from $S_T$ with $S\oplus S_T=[w_1\oplus v_1, w_2\oplus v_2,\dots,w_n\oplus v_n]$. Similarly, the words in aspect terms are appended with
their corresponding representations in $S_T$. In other words, to achieve our transfer learning goal we simply replace $S$ with $S\oplus S_T$ in Equations \ref{eq:alsa1}, \ref{eq:alsa2}. Finally, the overall ALSA algorithm can be stated as

\begin{flalign}
    S_T&=\BiGRU_{AE}(S), \label{eq:alsa0_T}\\
    A_i&=(S\oplus S_T)[h_i:k_i], 0\leq h_i\leq k_i \leq n, \label{eq:alsa2_T}\\
    l_i&=\ALSA_{\ast}((S\oplus S_T), A_i), \label{eq:alsa1_T}
\end{flalign}

where $S\in \mathbb{R}^{n\times d}$,  $S_T\in \mathbb{R}^{n\times D_T}$, 
$S\oplus S_T\in \mathbb{R}^{n\times (d+D_T)}$, $A_i\in \mathbb{R}^{(k_i-h_i+1)\times (d+D_T)}$, and $l_i\in \{0,1,2\}$ ($0,1, \text{and } 2$ stand for \emph{positive}, \emph{negative}, and \emph{neutral} sentiment, respectively).

\section{Experimental Settings}
\label{sec:experimental-settings}
We compare our AE assisted ALSA approach with the baseline ALSA models extensively.

\subsection{Dataset Details}

We evaluate our method on SemEval-2014 Task 4\footnote{\url{http://alt.qcri.org/semeval2014/task4/}} dataset, containing samples from two domains --- Laptop and Restaurant. Each domain consists of over 3K customer reviews of restaurants and laptops in English. \ref{tab:dataset-dist} shows the distribution of samples over two domains and two partitions --- training and test. Further, \ref{tab:dataset-asp} shows the number samples whose aspect share source sentence with at least one other aspect, denoted as MA (multi-aspect), and number of samples that have unique source sentences, denoted as SA (single-aspect). Both domain contain information relevant to the two subtasks --- AE and ALSA.

\begin{table}[h]
      \caption{Distribution of the samples by domain and class labels in SemEval 2014 dataset.}
      \begin{tabular}{c|cc|cc|cc}
      \hline
      \multirow{2}{*}{Domain} & \multicolumn{2}{c|}{Positive} & \multicolumn{2}{c|}{Negative} & \multicolumn{2}{c}{Neutral}\\
     \cline{2-7} & Train & Test & Train & Test & Train & Test\\
      \hline
      \hline
      Restaurant & 2,164 & 728 & 807 & 196 & 637 & 196\\
      Laptop & 994 & 341 & 870 & 128 & 464 & 169\\
      \hline
      \end{tabular}
    \label{tab:dataset-dist}
\end{table}

We transfer the embeddings learnt from AE, namely $S_T$, to the following ALSA methods, as per 
Equations \ref{eq:alsa0_T}, \ref{eq:alsa2_T}, \ref{eq:alsa1_T}. These ALSA methods also serve as the baselines that we compare our 
transfer learning scheme against.

\paragraph{Majority classifier} assigns the sentiment polarity with most samples in the training set to the all the samples in the test set.

\paragraph{TC-LSTM~\cite{tang-etal-2016-effective}} considers preceding ($S_L=[w_1,w_2,\dots,w_k]$) and succeeding ($S_R=[w_{k+p+1},w_{k+p+2},\dots,w_n]$) words to the target-aspect term as left and right context, respectively, where $p$ is the target-aspect term length. Each word in $S_L$ and $S_R$ is appended with the aspect-term representation, that is the mean of $[w_{k+1}, w_{k+2},\dots,w_{k+p}]$. Two distinct LSTMs $\text{LSTM}_L$ and $\text{LSTM}_R$ are used to model left and right context. $\text{LSTM}_L$ takes input from left to right along aspect-appended $S_L$ and $\text{LSTM}_R$ accepts input from right to left along aspect-appended $S_R$. The final output of $\text{LSTM}_L$ and $\text{LSTM}_R$ are concatenated to form the target-aspect-specific sentence representation, which is fed to a softmax layer to classify sentiment polarity.

\paragraph{ATAE~\cite{wang-etal-2016-attention}} employs attention mechanism to amplify parts of the sentence that are relevant to the task. It appends the aspect representation $a$ to the representation of each word in the sentence $S$, resulting [$w_1\oplus a, w_2\oplus a,\dots,w_n\oplus a$]. This is fed to an LSTM yielding outputs $H$=[$h_1, ..., h_n$]. Each $h_i$ is again appended to aspect representation $a$ and fed to an attention layer to obtain
relevance score $\alpha$ on each word in $S$. $H$ is pooled using $\alpha$ as weight to achieve target-aspect-specific sentence representation, which is fed to a softmax layer to classify sentiment polarity. \label{ATAE_Para}

\paragraph{IAN~\cite{ma2017interactive}} uses two different LSTMs $\text{LSTM}_A$ and $\text{LSTM}_S$ for aspect term and sentence encoding, respectively. The output of the LSTMs $H_A$ and $H_S$ are max-pooled to obtain $h_s$ and $h_a$, respectively. Interaction between aspect and sentence and vice versa is obtained by attending over $H_A$ with respect to $h_s$ and $H_S$ with respect to $h_a$, respectively. Pooling with the attention scores results final aspect and sentence representation that are concatenated together and fed to a softmax classifier.

\paragraph{Multi-task} learning aims to improve overall performance of related tasks when trained together \cite{zhang2017survey}. We adopt hard-parameter sharing strategy~\cite{liu2016recurrent, liu2019multi, yang2016multi} wherein multiple tasks share certain hidden layers, however, they have separate output layers. In our multitask setting, we perform aspect extraction and aspect level sentiment analysis in the same network by sharing the weights in the initial BiGRU layer that is tasked for sentence encoding in the network. Specifically, given a sentence representation $S = [w_1,...,w_n]$ is fed to a BiGRU to obtain $S_c$ that is common to both the tasks AE and ALSA. \ref{eq:AE_1} describes the remaining procedure for AE where $S$ is replaced with $S_c$. For ALSA, we choose ATAE architecture.
\begin{table}
\centering
    \caption{Distribution of the samples by the appearance of single aspect (SA) and multiple aspects (MA) in the source sentence in SemEval 2014 dataset.}
    \begin{tabular}{c|cc|cc}
      \hline
      \multirow{2}{*}{Domain} & \multicolumn{2}{c|}{Train} & \multicolumn{2}{c}{Test}\\
      \cline{2-5} & SA & MA & SA & MA \\
      \hline
      \hline
      Restaurant & 1,063 & 2,545 & 302 & 818 \\
      Laptop & 957 & 1,371 & 269 & 369 \\
      \hline
    \end{tabular}
    \label{tab:dataset-asp}
\end{table}

\subsection{Transfer Learning on ALSA Models}

\begin{figure*}[h]
    \begin{center}
    \includegraphics[width=0.8\linewidth]{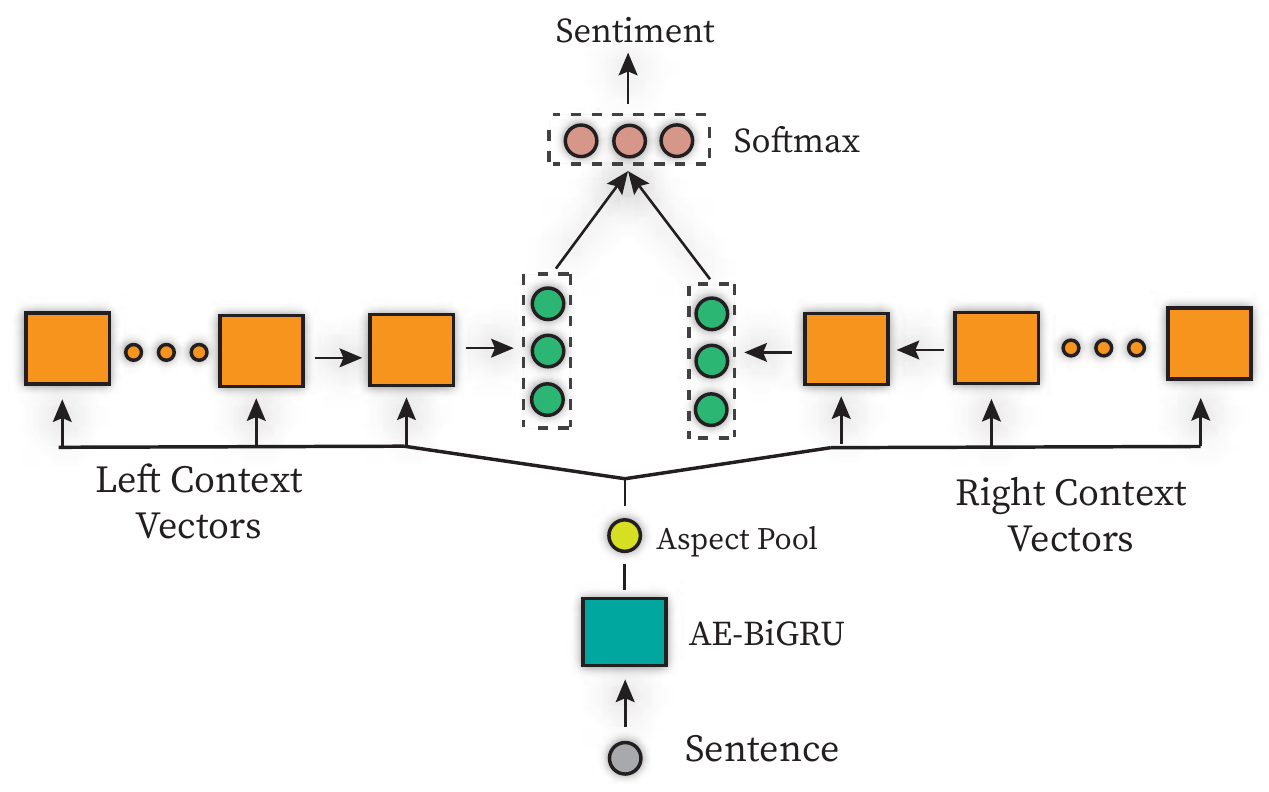}
    \caption{Transfer learning on TC-LSTM.}
    \label{fig:TC_LSTM}
    \end{center}
    \end{figure*}

\begin{figure*}[h]
    \begin{center}
    \includegraphics[width=0.8\linewidth]{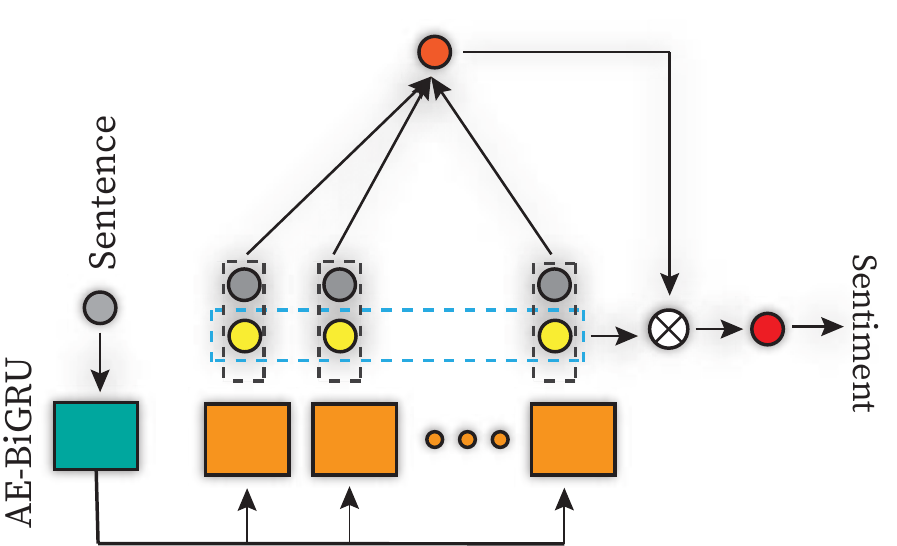}
    \caption{Transfer learning on ATAE.}
    \label{fig:ATAE}
    \end{center}
    \end{figure*}

\begin{figure*}[h]
    \begin{center}
    \includegraphics[width=0.8\linewidth]{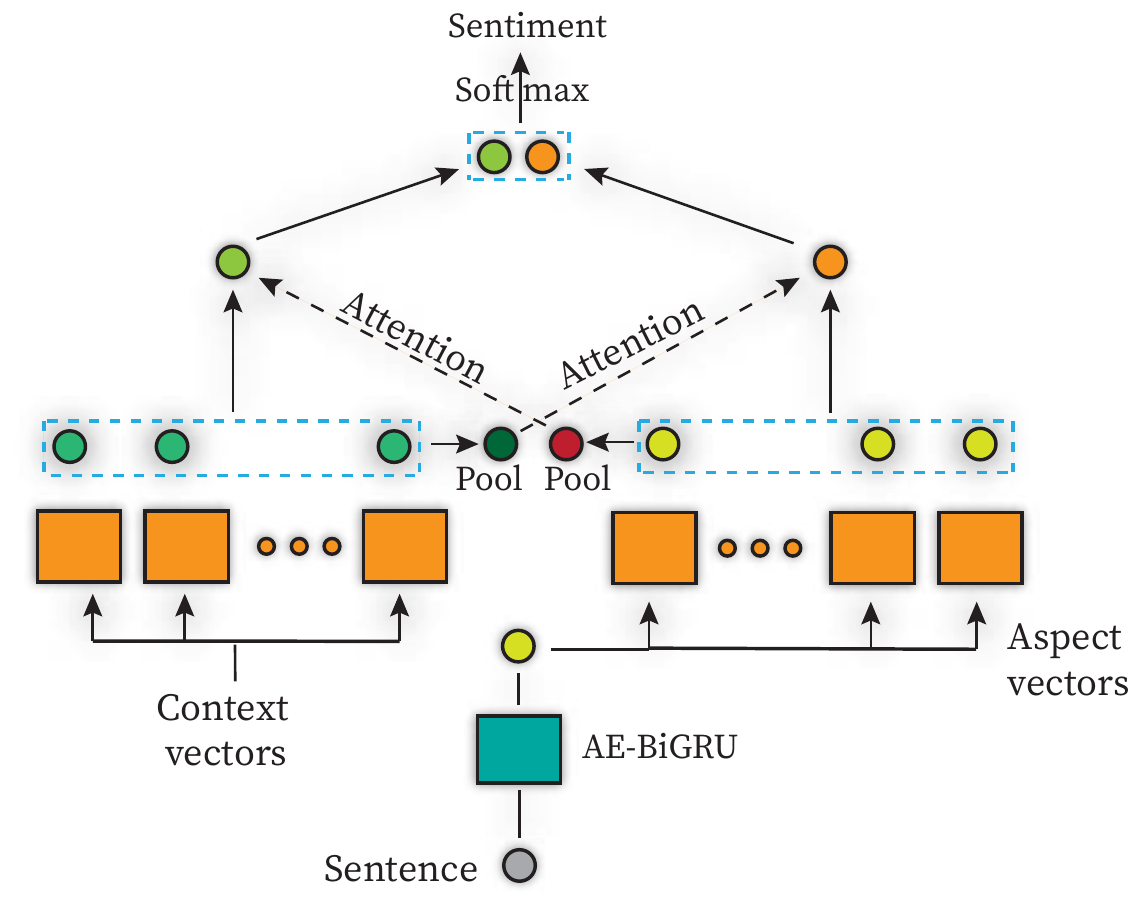}
    \caption{Transfer learning on IAN.}
    \label{fig:IAN}
    \end{center}
    \end{figure*}

We transfer the word representations learnt in AE to ALSA using Equations \ref{eq:alsa0_T}, \ref{eq:alsa2_T}, \ref{eq:alsa1_T}. The transfer learning variants of TC-LSTM, ATAE, and IAN are denoted with TC-LSTM-T~(\ref{fig:TC_LSTM}), ATAE-T~(\ref{fig:ATAE}), and IAN-T~(\ref{fig:IAN}), respectively.
The input word embeddings of each such model come from $S\oplus S_T$, instead of $S$.

To gauge the impact of the knowledge from AE, defined by $S_T$ in \ref{eq:alsa0_T}
, we substitute the relationship-aware word representations $v_i$ with randomly generated
vectors $r_i\in \mathbb{R}^{D_T}$ whose each element is sampled from $\mathcal{N}(0,1)$.
Such variants are named TC-LSTM-R, ATAE-R, and IAN-R after their respective models
TC-LSTM, ATAE, and IAN.
Moreover, we evaluate the generality of the embeddings from AE, $S_T$, using
cross-domain knowledge transfer.

We used stochastic gradient descent-based Adam~\cite{DBLP:journals/corr/KingmaB14} optimizer to train the models. 
As training loss, all the ALSA models use categorical cross-entropy. 
Hyper-parameter tuning was done using grid search. \ref{tab:hyper-param} shows the optimal hyper-parameters that we found for different models.

\begin{table}[h]
    \caption{Optimal hyper-parameters; $lr$ stands for learning rate; 
    $\lambda$ stands for L2-regularization weight.
    }
    \begin{tabular}{c|ccc|ccc}
      \hline
      \multirow{2}{*}{Model} & \multicolumn{3}{c|}{Laptop} & \multicolumn{3}{c}{Restaurant} \\
      \cline{2-7} & $lr$ & $D_T$ & $\lambda$ & $lr$ & $D_T$ & $\lambda$\\
      \hline
      \hline
      AE & 0.001 & 64 & -- & 0.001 & 64 & -- \\
      TC-LSTM & 0.002 & -- & 0.001 & 0.002 & -- & $1e^{-6}$\\
      ATAE & 0.001 & -- & 0.001 & 0.002 & -- &$1e^{-5}$\\
      IAN & 0.002 & -- & $1e^{-5}$ & 0.002 & -- & $1e^{-5}$\\
      \hline
      TC-LSTM-T & 0.002 & -- & $1e^{-6}$ & 0.002 & -- & $1e^{-6}$\\
      ATAE-T & 0.002 & -- & 0.001 & 0.002 & -- & 0.001\\
      IAN-T & 0.001 & -- & $1e^{-6}$ & 0.002 & -- & $1e^{-5}$\\
      \hline
    \end{tabular}
    \label{tab:hyper-param}
\end{table}
\section{Results and Discussion}
\label{sec:results}

\begin{table}
\centering
    \caption{Macro F1-scores for different ALSA models; multi-task model jointly performs AE and ALSA.}
    \begin{tabular}{c|c|c}
    \hline
    Model &Laptop &Restaurant \\
    \hline
    \hline
    Majority & 23.22 & 26.26 \\
    \hline
    TC-LSTM &61.96 &66.00\\
    ATAE &62.34 &65.24\\
    IAN &64.86 &66.41\\
    \hline
    TC-LSTM-R &61.41 &65.05\\
    ATAE-R &62.32 &63.12\\
    IAN-R &58.58 &59.96\\
    \hline
    Multi-Task &56.43 &51.17\\
    \hline
    TC-LSTM-T & 63.31 & 66.79 \\
    ATAE-T & {\bf 66.09} & 66.17 \\
    IAN-T & 65.82 & {\bf 67.81} \\
    \hline
    \end{tabular}
 \label{tab:baselines}
 \end{table}
 
\subsection{Overall Comparison}
 
It is evident from Table \ref{tab:baselines} that all three baseline ALSA models, namely TC-LSTM, ATAE, and IAN, are surpassed by their corresponding transfer learning-based counterparts, TC-LSTM-T, ATAE-T, and IAN-T, respectively, by 1.53\% on average across two domains. 

Interestingly, ATAE-T outperforms IAN-T by a small margin on laptop domain, in contrast with the restaurant domain and the setup without transfer learning. Since, IAN is a more complex model than ATAE, with two LSTMs and two attention layers for sentence and aspect interacting with each other, it is better capable of learning the relationships between words compared to ATAE, with enough data. These relationships are crucial for identifying the words related to the target aspect, which influences the identification of the sentiment-bearing words for classification. In transfer learning setup, the embeddings from AE, $S_T$, more directly convey these relationships to ATAE, improving its performance in ATAE-T form. However, for IAN-T, being more complex, requires more training data to fully utilize this extra information and to resolve conflict with its own features, if any. Training set is lacking in laptop domain as compared to restaurant domain (Table \ref{tab:dataset-dist}). As such, ATAE, being simpler, makes slightly better use of the transferred knowledge from AE.

TC-LSTM-R, ATAE-R, and IAN-R models, where the knowledge from AE is replaced with
random noise, underperform compared to their regular counterparts. This is expected since
the models basically learn to ignore this noise, but cannot ignore completely. The 
performance drop is minimal for TC-LSTM-R and ATAE-R in general. However, the drop is 
massive for IAN-R. This is again the consequence of the complexity of IAN, which entails
more difficulty in ignoring the noise.

The multi-task model performs the worst among all. We assume this is due to AE
and ALSA tasks being structurally different from each other. There are features that
are required exclusively to AE or ALSA, not both. Since, in the multi-task setup AE and
ALSA share some initial layers processing the input sentence, these two tasks compete
each other for features within those layer. This leads to poorer performance on both of
tasks, compared to task-specific models.

\subsection{In-Domain vs Cross-Domain Knowledge Transfer}
Ideally, in NLP tasks the cross-domain performance tends to be worse than the in-domain performance~\cite{DBLP:conf/acl/PlankR18,elsahar-galle-2019-annotate}.
If we train our AE model on the laptop domain and transfer that knowledge to the ALSA model of the restaurant domain and vice versa, can the performance drop because of this cross-domain knowledge transfer? The result is interesting as it can be seen in \ref{tab:inter-intra}.

\begin{table}[h]
\centering
 \caption{Macro F1-scores for in- and cross-domain transfer learning; column headers indicate the domain of aspect extraction and the row headers represent the domain of ALSA.}
 \newcommand\stack[2]{$\stackrel{\displaystyle \text{#1}}{\text{#2}}$}
     \begin{tabular}{c|c|c|c}
     \hline
     \multirow{2}{*}{\stack{ALSA}{Domain}}&\multirow{2}{*}{Model}&\multicolumn{2}{c}{AE Domain} \\
     \cline{3-4}
      & & Laptop & Restaurant\\
     \hline
     \hline
     \multirow{3}{*}{Laptop} &TC-LSTM-T & {\bf 63.31} & 62.94 \\
     & ATAE-T & {\bf 66.09} & 64.41 \\
     & IAN-T & 65.82 & {\bf 65.95} \\
     \cline{2-4}
     \multirow{3}{*}{Restaurant} & TC-LSTM-T & {\bf 68.24} & 66.79 \\
     & ATAE-T & {\bf 66.77} & 66.17 \\
     & IAN-T & 66.86 & {\bf 67.81} \\
    \hline
    \end{tabular}
 \label{tab:inter-intra}
 \end{table}
 
In Table \ref{tab:inter-intra}, we depict that the knowledge from the AE model trained on the laptop domain boosts the performance of TC-LSTM and ATAE ALSA models in both domains. However, the cross-domain performance of the AE model trained in the restaurant domain is worse than its in-domain counterparts for two out of the three baselines. AE is a task that greatly relies on the syntactic structure of the input~\cite{poria2016aspect}. According to them, the dependency-based syntactic rules can give decent accuracy on the AE task without any need to semantically modelling the data. Hence, as long as the training set has consistently grammatical sentences with rich annotations, the AE system can perform well on the other domains too which we think one of the key reasons why the cross-domain setting in our experiments showing comparable results to their in-domain counterparts.

\subsection{Single Aspect vs Multiple Aspect Case}

\begin{table}[h]
  \centering
    \caption{Macro F1-scores for samples with single aspect (SA) and multiple aspects (MA). }
    \begin{tabular}{c|cc|cc}
      \hline
      \multirow{2}{*}{Model} & \multicolumn{2}{c|}{Laptop} & \multicolumn{2}{c}{Restaurant} \\
      \cline{2-5} 
       & SA & MA & SA & MA\\
      \hline
      \hline
      TC-LSTM & 60.07 & 63.59 & 69.73 & 63.27 \\
      ATAE & 60.96 & 63.73 & 64.17 & 63.67 \\
      IAN &61.24 &64.01 &63.73 &68.11 \\
      \hline
      TC-LSTM-T & 60.08 & 65.87 & 70.84 & 67.21 \\
      ATAE-T & {\bf 68.68} & {\bf 67.27} & {\bf 70.85} & 64.37 \\
      IAN-T & 64.41 & 66.13 & 70.18 & {\bf 68.16} \\
      \hline
    \end{tabular}
    \label{tab:single-multi}
\end{table}

Following Table \ref{tab:single-multi}, on both domains and single- and multi-aspect cases, the models with external AE knowledge (*-T) significantly outperforms their standalone counterparts, as expected, by 3.15\% overall. On laptop domain, for both single- and multi- aspect cases ATAE-T performs the best. We
surmise this is due to ATAE being simple enough to utilize the knowledge from AE with 
small number of samples. On restaurant domain, however, ATAE-T marginally surpasses IAN-T and TC-LSTM-T for single-aspect case. For multi-aspect case, IAN-T performs the best
due to IAN being more capable for multi-aspect cases than the others and restaurant domain having almost twice as much multi-aspect training samples than laptop domain (Table \ref{tab:dataset-asp}).

\subsection{Class-Wise Comparison}

\begin{table}[h]
    \centering
    \caption{Label-wise macro F1-scores of ATAE.}
        \begin{tabular}{c|c|c|c}
        \hline
    Class    & Model  & Laptop & Restaurant \\ \hline \hline
    \multirow{2}{*}{Positive} & ATAE   & 81.75  & 85.60      \\ 
             & ATAE-T & {\bf 83.69}  & {\bf 87.69}      \\ \hline \hline
    \multirow{2}{*}{Negative} & ATAE   & 58.33  & 60.00      \\ 
             & ATAE-T & {\bf 64.72}  & {\bf 68.01}      \\ \hline \hline
    \multirow{2}{*}{Neutral}  & ATAE   & 48.39  & {\bf 45.64}      \\ 
             & ATAE-T & {\bf 57.14}  & 42.95      \\ \hline     
        \end{tabular}
    \label{tab:label_dist}
\end{table}

As per Table \ref{tab:label_dist}, as expected, ATAE-T outperforms ATAE on all domain and class combinations but on restaurant domain for neutral class. ATAE performs better here. This aberration might be a result of smaller quantity neutral samples in both domain, which made the performance unstable. We suppose that with added neutral samples the performance of ATAE-T would surpass that of ATAE.

\subsection{Case Study}
 
\begin{figure}[h]
    \centering
    \includegraphics[width=\linewidth]{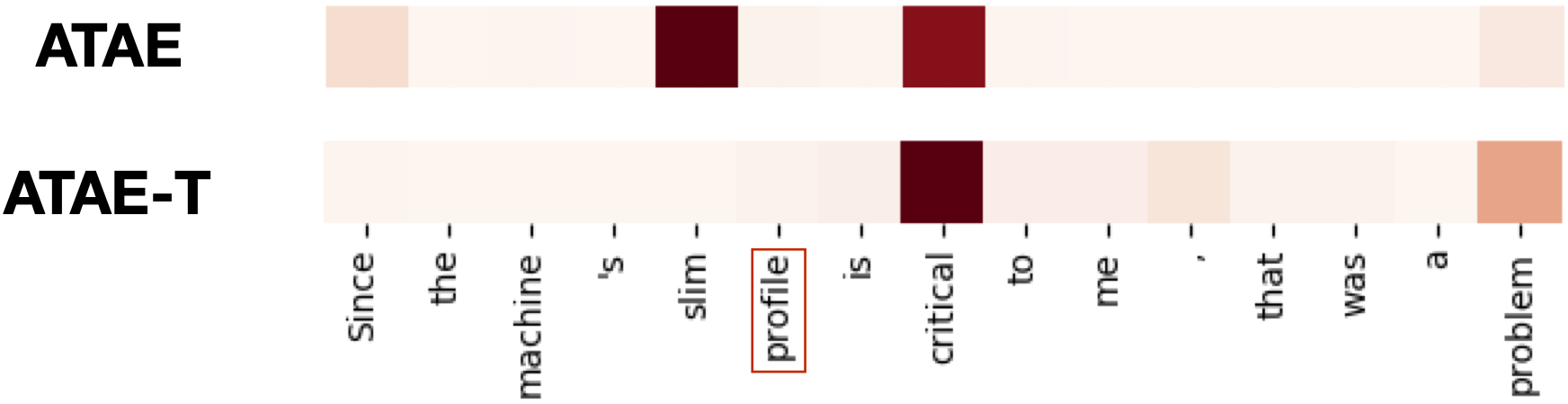}
    \caption{Comparison between attention vectors from ATAE and ATAE-T for aspect `\emph{profile}'.}
    \label{fig:single-label}
\end{figure}

Figure \ref{fig:single-label} illustrates a case where ATAE and ATAE-T classifies aspect `\emph{profile}' in the sentence ``\emph{Since the machine's slim profile is critical to me, that was a problem.}''. Here, ATAE misclassifies the sentiment as \emph{neutral} due to its focus on the words `slim' and `critical', both of which carry \emph{neutral} sentiment. ATAE-T, on the other hand, focuses on `critical' and `problem' where the latter one has \emph{negative} sentiment, which leads to correct classification as \emph{negative}. We posit that this wrong focus by ATAE is caused by its lack of understanding on relation between target aspect and context words. Due to the sentiment word `problem' appearing far away from the target, ATAE was unable to make this association on its own. ATAE-T overcomes this deficiency using the knowledge from AE. We encountered similar instances across the test set.

\begin{figure}
    \centering
    \includegraphics[width=\linewidth]{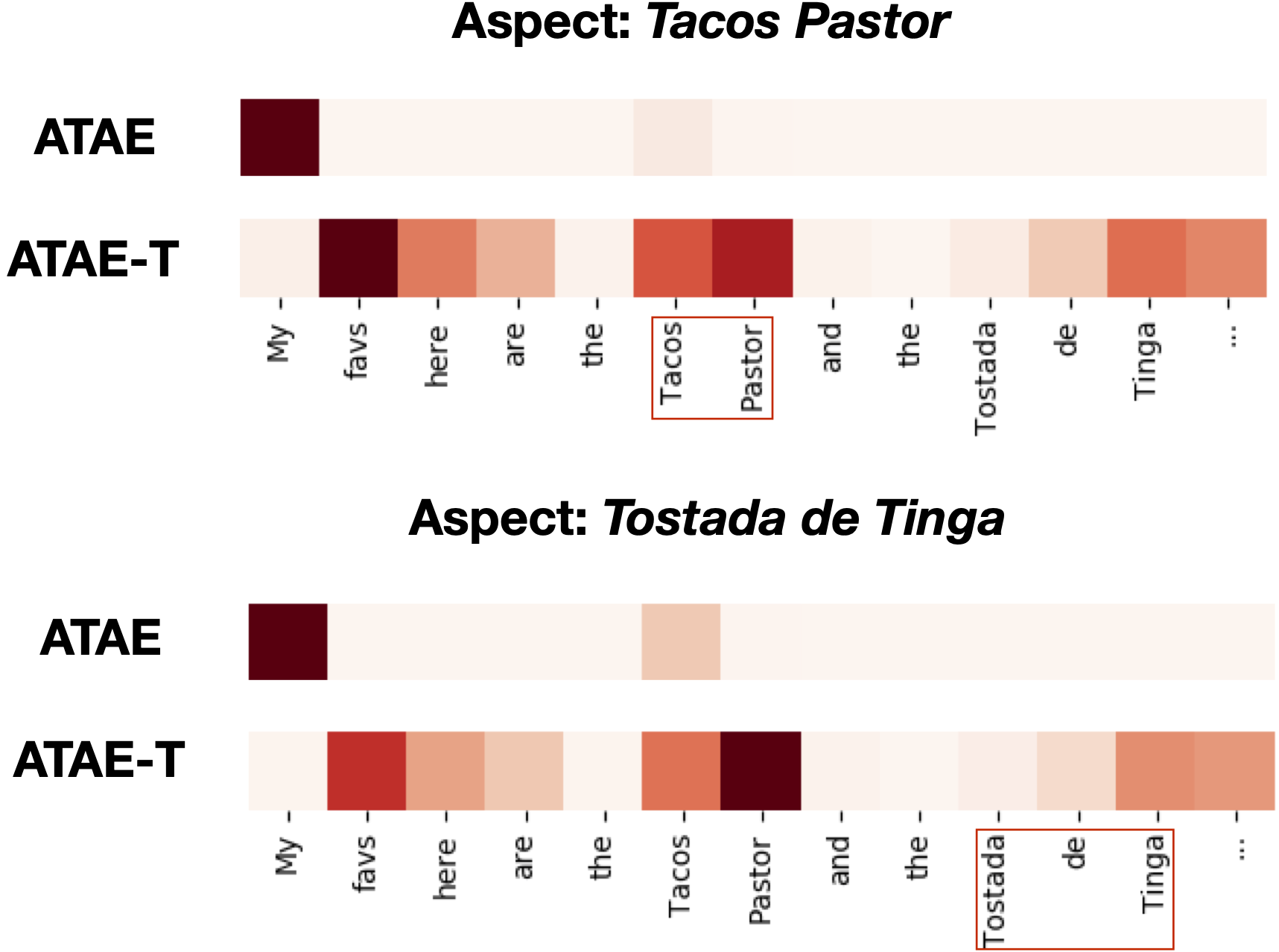}
    \caption{Comparison between attention vectors from ATAE and ATAE-T for two aspects from a single sentence.}
    \label{fig:multi-label}
\end{figure}

This lack of syntactic understanding without AE is evident for multi-aspect scenario as well, as shown in Figure \ref{fig:multi-label}. ATAE misclassifies both of the aspects `Tacos Pastor' and `Tostada de Tinga', as in both cases it focuses on a \emph{neutral} word `my'. ATAE-T, however, leverages knowledge from AE to focus on the appropriate \emph{positive} word `favs' to make accurate classification as \emph{positive}. Such instances were frequent within our observation. 

\subsection{Error Analysis}

\begin{figure}
    \centering
    \includegraphics[width=\linewidth]{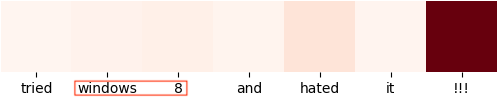}
    \caption{Misclassification of aspect `\emph{windows 8}' due to wrong focus by ATAE-T.}
    \label{fig:error-single-label}
\end{figure}

A prevalent trend we observed for both ATAE and ATAE-T is the strong focus on exclamation marks in the sentence. In Figure \ref{fig:error-single-label}, clearly, the sentiment of aspect `windows 8' is \emph{negative} due to the presence of the word `hated'. ATAE-T, unfortunately, focuses on `!!!', which is usually a sign of excitement and often times \emph{positive}, leading to \emph{positive} prediction. To mitigate such errors we need more samples with `!' punctuation used in various contexts, both \emph{positive} and \emph{negative}. Further, embeddings trained on twitter data can be used where `!' is prevalent. We leave this to our future work.

\begin{figure}
    \centering
    \includegraphics[width=\linewidth]{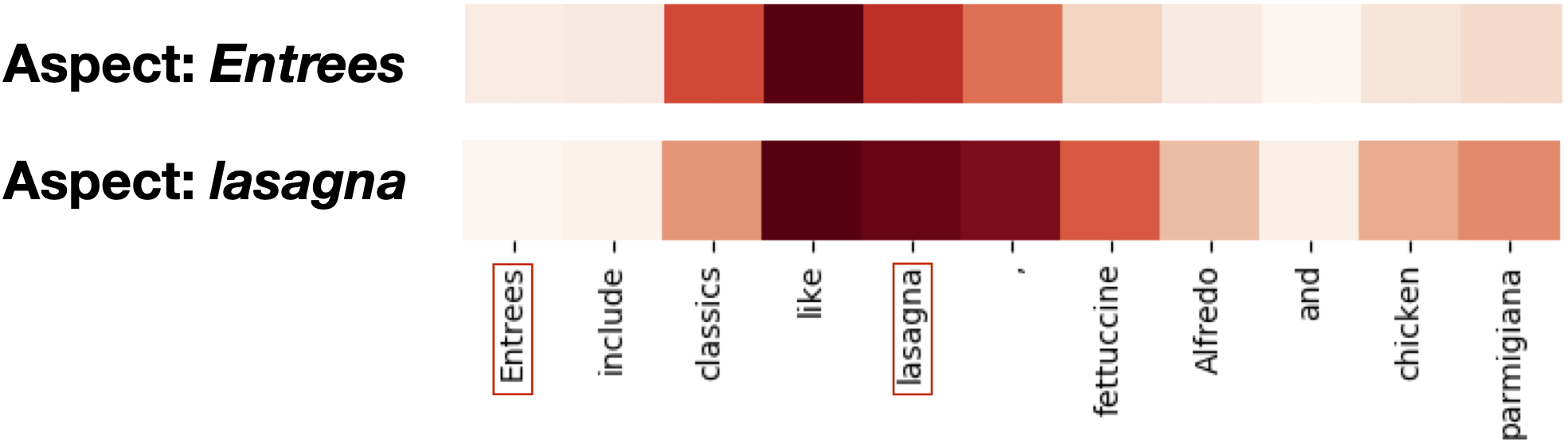}
    \caption{Misclassification of multiple aspects due to wrong focus and interpretation by ATAE-T.}
    \label{fig:error-multi-label}
\end{figure}

Figure \ref{fig:error-multi-label} shows a case wherein ATAE-T misattributes the sentiment of both aspects `Entrees' and `lasagna' to the word `like', which is often a \emph{positive} word, leading to misclassification as \emph{positive}. However, clearly in this context --- ``\emph{Entrees include classics like lasagna, fettuccine Alfredo and chicken parmigiana.}'' --- `like' carries no emotion (\emph{neutral}). Further, there are no sentimentally-charged words in the sentence. However, our model fails to learn this due to relatively small number of \emph{neutral} samples. To compensate for this small count, we could use word-sense disambiguation to indicate the true sense of words as `like', given the context. Even, context-sensitive embeddings like BERT~\cite{devlin-etal-2019-bert} can be utilized.
 
\section{Conclusion}
\label{sec:conclusions}

We experimentally show that aspect extraction (AE) can substantially aid aspect-level sentiment analysis (ALSA) by passing the gained knowledge in AE to ALSA. Further, we show this knowledge transfer across different domains is quite viable. This domain-invariant knowledge transfer allows AE to be trained on larger general domain datasets. This improvement translates down to single-aspect and multi-aspect scenarios and to each class also. However, there remains ample room for improvement, as our approach struggles where the semantics is ambiguous or the dataset is limited. Further, we surmise leveraging external knowledge --- specifically, knowledge graphs like ConceptNet~\cite{conceptnet} --- has strong possibility of improving cross-domain performance by connecting domain-specific aspects through the edges in knowledge graph. We plan to explore these avenues of ALSA in the future.

\begin{acknowledgements}
This research is supported by A*STAR under its RIE 2020 Advanced Manufacturing and Engineering (AME) programmatic grant, Award No. -  A19E2b0098, Project name - K-EMERGE: Knowledge Extraction, Modelling, and Explainable Reasoning for General Expertise.
\end{acknowledgements}

%
\section*{Conflict of interest}

The authors declare that they have no conflict of interest.

\bibliographystyle{spphys}       
\bibliography{NCA}   

\end{document}